\documentclass[letterpaper, 10 pt, conference]{ieeeconf}  

\IEEEoverridecommandlockouts                              

\title{\LARGE \bf
Panoramic Direct LiDAR-assisted Visual Odometry
}

\author{Qirui~Hu$^{1\dag}$, Zikang~Yuan$^{2\dag}$, Tianle~Xu$^{3\dag}$, Xiaoxiang~Wang$^{3}$, Jinni~Geng$^{4}$ and Xin~Yang$^{3*}$ 
\thanks{$^{1}$Qirui~Hu is with Li Siguang School, China University of Geosciences, Wuhan, 430074, China. E-mail: {\tt\small qiruihu@cug.edu.cn}}%
\thanks{$^{2}$Zikang~Yuan is with Institute of Artificial Intelligence, Huazhong University of Science and Technology, Wuhan, 430074, China. E-mail: {\tt\small 403525990@qq.com}}%
\thanks{$^{3}$Tianle~Xu, Xiaoxiang Wang and Xin Yang are with School of Electronic Information and Communications, Huazhong University of Science and Technology, Wuhan, 430074, China. E-mail: {\tt\small \{tianlexu, xiaoxiang0012, xinyang2014\}@hust.edu.cn}}%
\thanks{$^{4}$Jinni~Geng is with School of Optoelectronic Engineering, Xidian University, Xian, 710126, China. E-mail: {\tt\small haofan0828@163.com}}
\thanks{$^{\dag}$Qirui~Hu, $^{\dag}$Zikang~Yuan and $^{\dag}$Tianle~Xu contributed  equally  to  this  work. * represents the corresponding author.}
}

\usepackage{cite}
\usepackage{amsmath,amssymb,amsfonts}
\usepackage{algorithmic}
\usepackage{graphicx}
\usepackage{textcomp}
\usepackage{xcolor}
\usepackage{multirow}
\usepackage{threeparttable}
\usepackage{array}
\usepackage{booktabs}
\usepackage{multirow}
\usepackage{amssymb}
\usepackage{amsfonts}
\usepackage{amsbsy}

\begin{document}

\maketitle
\thispagestyle{empty}
\pagestyle{empty}

\begin{abstract}

Enhancing visual odometry by exploiting sparse depth measurements from LiDAR is a promising solution for improving tracking accuracy of an odometry. Most existing works utilize a monocular pinhole camera, yet could suffer from poor robustness due to less available information from limited field-of-view (FOV). This paper proposes a panoramic direct LiDAR-assisted visual odometry, which fully associates the 360-degree FOV LiDAR points with the 360-degree FOV panoramic image datas. 360-degree FOV panoramic images can provide more available information, which can compensate inaccurate pose estimation caused by insufficient texture or motion blur from a single view. In addition to constraints between a specific view at different times, constraints can also be built between different views at the same moment. Experimental results on public datasets demonstrate the benefit of our panoramic direct LiDAR-assisted visual odometry to state-of-the-art approaches.

\end{abstract}

\section{Introduction}
\label{Introduction}

\begin{figure}
	\begin{center}
		\includegraphics[width=1\linewidth]{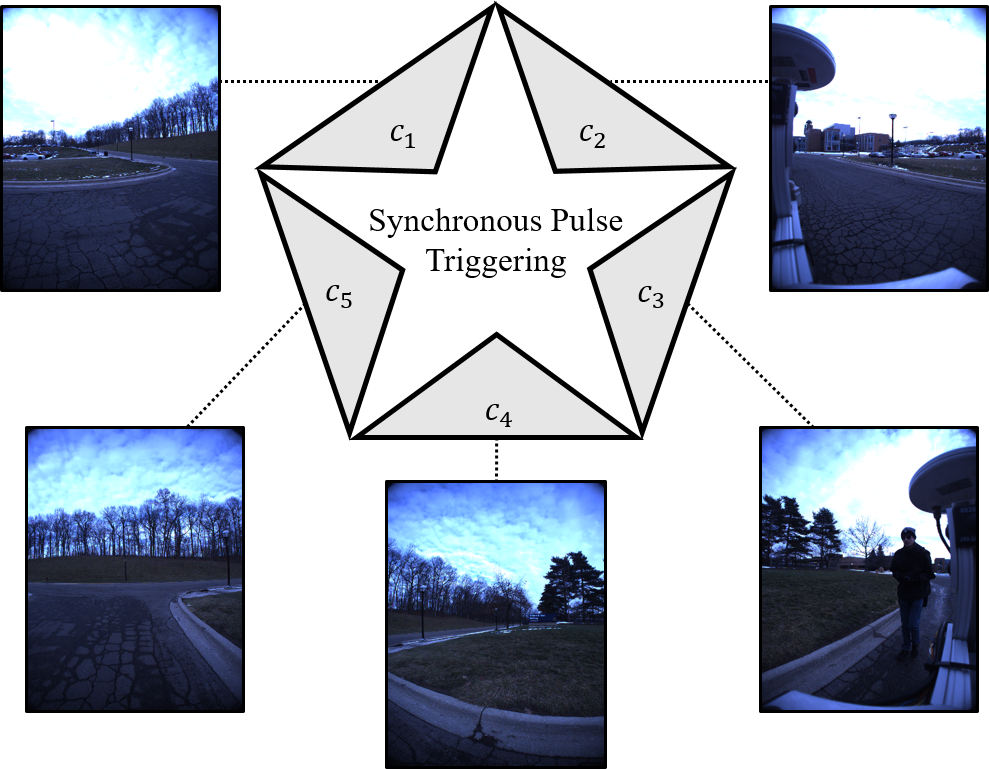}
		\caption{Illustration of a panoramic camera, which collects 360-degree FOV image datas through multiple synchronized surround-view cameras. Compared to utilizing rigidly coupled multi-camera systems to achieve 360-degree FOV, the panoramic camera has the advantage of not requiring online extrinsic calibration and temporal synchronization between multiple cameras.
		}
		\label{fig1}
	\end{center}
\end{figure}

Using 3D light detection and ranging (LiDAR) information to assist visual odometry (VO), i.e., LiDAR-assisted VO, has attracted increasing interests recently due to the ability of achieving accurate pose estimation in outdoor environments. Several notable LiDAR-assisted VO systems \cite{zhang2017real, graeter2018limo, huang2020lidar, shin2020dvl, yuan2023sdv, yuan2024sr} fuse measurements from a monocular pinhole camera and a 360-degree field-of-view (FOV) LiDAR to perform pose estimation. These approaches suffer from poor robustness if the single view appears motion blur or insufficient texture. Compared to the monocular camera, the panoramic camera provides a 360-degree FOV image data (as shown in Fig. \ref{fig1}), which can more fully associate images with LiDAR points and hold great potential to enhance the robustness of LiDAR-assisted VO. Compared to utilizing rigidly coupled multi-camera systems to achieve 360-degree FOV, the panoramic camera has the advantage of not requiring online extrinsic calibration and temporal synchronization between multiple cameras.

Despite a panoramic camera can provide 360-degree FOV through multiple synchronized surround-view cameras, each camera has a low frame rate and a very limited horizontal FOV. When the camera moves quickly, there is little overlap between consecutive frames from the same perspective. The limited overlap results in less information available for constructing constraints, which in turn leads to poor accuracy in pose estimation.

In this paper, we propose a panoramic direct LiDAR-assisted visual odometry, for accurate and robust pose estimation in outdoor environments. By using a panoramic camera with 360-degree FOV, we can fully associate the 360-degree FOV LiDAR points with the 360-degree FOV panoramic image datas. To address the issue of limited overlap between consecutive frames from a specifica view, we allow the view from a historical moment to construct constraints with any view from the current moment. Compared to using images from the same view at different times to build constraints individually, our approach can utilize available image information from a broader area, thereby constructing more constraints and enhancing the accuracy of pose estimation. Experimental results on two public datasets, i.e., $nclt$ \cite{carlevaris2016university} and $ijrr$ \cite{pandey2011ford}, demonstrate the benefit of associating 360-degree FOV LiDAR points with 360-degree FOV panoramic image datas, and the effectiveness of constructing constraints between different views.

To summarize, the main contributions of this work are two folds: (1) To the best of our knowledge, we firstly propose a LiDAR-assisted VO framework that uses a panoramic camera and a 3D LiDAR. (2) We have released the source code of our system to benefit the development of the community\footnote{https://github.com/ZikangYuan/panoramic\_lidar\_dso}.

The rest of this paper is structured as follows. Sec. \ref{Related Work} reviews the relevant literatures. Sec. \ref{Preliminaries} provides preliminaries. Secs. \ref{System Overview} and \ref{System Details} presents system overview and details, followed by experimental evaluation in Sec. \ref{Experiments}. Sec. \ref{Conclusion} concludes the paper.

\section{Related Work}
\label{Related Work}

In this section, we review the related work about existing wide FOV camera-based VO, multiple cameras-based VO, panoramic VO and monocular LiDAR-assisted VO.

\textbf{Wide FOV camera-based VO.} Zhang et. al. \cite{zhang2016benefit} conducted an in-depth investigation into the impact of camera FOV on the performance of VO, and pointed out that it is advantageous to use a large FOV camera (e.g., fisheye camera) for indoor scenes. Caruso et. al. \cite{caruso2015large} proposed a direct monocular SLAM method for omnidirectional or wide FOV fisheye cameras, which allow to observe and reconstruct a larger portion of the surrounding environment, and also make the system more robust to degenerate (rotation-only) motion. Matsuki et. al. \cite{matsuki2018omnidirectional} proposed a direct monocular visual odometry for omnidirectional cameras. \cite{matsuki2018omnidirectional} utilized the unified omnidirectional model as a projection function, which can be applied to fisheye cameras with a FOV well above 180 degrees. CubemapSLAM \cite{wang2019cubemapslam} presented a real-time feature-based SLAM system for fisheye cameras featured by a large FOV, which increase visual overlap between consecutive frames and capture more pixels belonging to the static parts
of the environment. PALVO \cite{chen2019palvo} applied panoramic annular lens to visual odometry, greatly increasing the robustness to rapid motion and dynamic scenarios.
 
\textbf{Multiple camera-based VO.} Liu et. al. \cite{liu2018towards} combined several stereo cameras to obtain 360-degree FOV image datas, and in tutn improved the robustness of VO. ROVO \cite{seok2019rovo} proposed a robust visual odometry system for a wide-baseline camera rig with wide FOV fisheye lenses, which provides full omnidirectional stereo observations of the environment. Wang et. al. \cite{wang2021research} proposed an Omni-directional SLAM based on a forward binocular camera and three monocular cameras at the direction of left, right and rear. Multicol-SLAM \cite{urban2016multicol} extended and improved upon ORB-SLAM \cite{mur2015orb} to make it applicable to arbitrary, rigidly coupled multi-camera systems using the MultiCol model \cite{urban2017lafida}.

\textbf{Panoramic VO.} PAN-SLAM \cite{ji2020panoramic} proposed a panoramic feature-based SLAM system, which achieved accurate and robust camera localization and sparse map reconstruction in both small-scale indoor and large-scale outdoor environments. Jiang et. al. \cite{jiang2021panoramic} integrated the measurements form inertial measurement unit (IMU) and wheel encoder to PAN-SLAM in a tightly coupled manner. 360VO \cite{huang2022360vo} utilized a spherical camera model to process equirectangular images without rectification to attain omnidirectional perception. Different from most existing approaches, PVO \cite{lin2018pvo} directly used panoramic images without converting them to pinhole images, saving a lot of computing resources. In contrast to multiple camera-based VO, panoramic VO has the advantage of not requiring online extrinsic calibration and temporal synchronization between multiple cameras.

\textbf{LiDAR-assisted VO.} DEMO \cite{zhang2017real} proposed a feature-based LiDAR-assisted VO, which associates extracted 2D features with 3D points in order to assign depth value to each feature point (i.e., 2D-3D data association). LIMO \cite{graeter2018limo} added a loop closure module on the base of DEMO and utilizes semantic information to identify moving objects and reject outliers. Huang et. al. \cite{huang2020lidar} proposed to use line features in addition to point features for tracking and mapping, so as to improve the robustness to noises \cite{smith2006real}, large viewpoint changes \cite{gee2006real}, and motion blurs \cite{klein2008improving}. DVL-SLAM \cite{shin2020dvl} proposed to use the direct method for LiDAR-assisted visual tracking, aiming to avoid interpolation error in 2D-3D data association.

\section{Preliminaries}
\label{Preliminaries}

\subsection{Coordinate Systems}
\label{Coordinate Systems}

There are several types of panoramic cameras, and we specifically refer to the type that contains five surround-view cameras in this work (i.e., $c_1 \sim c_5$). We denote $(\cdot)^{c_i}$, $(\cdot)^b$, $(\cdot)^w$ as a 3D point in the $i_{th}$ camera coordinate, the body coordinate and the world coordinate respectively. The body coordinate is coinciding with $(\cdot)^{c_1}$, and $\mathbf{T}_b^{c_i}$ is the transformation form $(\cdot)^b$ to $(\cdot)^{c_i}$. The world coordinate system $(\cdot)^w$ coincides with the body coordinate system at the initial time.

\subsection{Camera Projection Model}
\label{Camera Projection Model}

Given a 3D point in $(\cdot)^w$ as $\mathbf{p}^{w}=(x, y, z)^{T} \in \mathbb{R}^{3}$ and its projection to a 2D image as $\mathbf{u}=(u, v)^{T} \in \mathbb{R}^{2}$, the camera projection model $\pi$: $\mathbb{R}^{3} \rightarrow \mathbb{R}^{2}$ is as follow:
\begin{equation}
	\label{equation1}
	\mathbf{u}=\pi\left(\mathbf{T}_{w}^{c} \mathbf{p}^{w}\right)
\end{equation}
where $\pi$ is determined by the $3 \times 3$ intrinsic camera parameters $\mathbf{K}$. 3D points in camera coordinates can be recovered from their 2D projections by the inverse projection model $\pi^{-1}$: $\pi$: $\mathbb{R}^{2} \rightarrow \mathbb{R}^{3}$:
\begin{equation}
	\label{equation2}
	\mathbf{p}^{c}=\pi^{-1}\left(\mathbf{u}, d_{\mathbf{u}}\right)
\end{equation}
where $d_{\mathbf{u}} \in \mathbb{R}$ represents the depth of 2D point $\mathbf{u}$ in the camera coordinate.

\section{System Overview}
\label{System Overview}

\begin{figure}
	\begin{center}
		\includegraphics[width=1\linewidth]{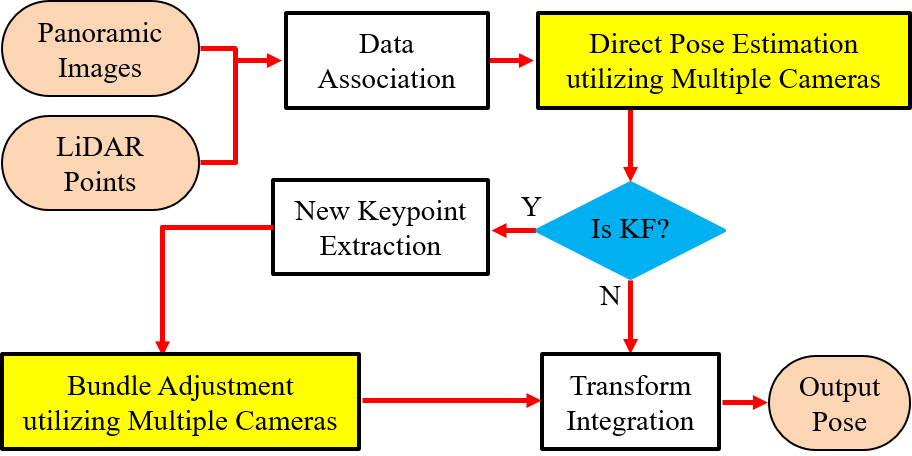}
		\caption{Overview of our system which consists of five main modules: data association, direct pose estimation utilizing multiple cameras, new keypoint extraction, bundle adjustment utilizing multiple cameras and transorm integration. The whole system is expanded from the monocular DSO \cite{engel2017direct}. We integrated the LIDAR observation into \cite{engel2017direct} and proposed corresponding tracking modules (labeled as yellow rectangles) to adapt to the characteristies of panoramic cameras.
		}
		\label{fig2}
	\end{center}
\end{figure}

Fig.\ref{fig2} illustrates the overview of our system which consists of five main modules: data association, direct pose estimation utilizing multiple cameras, new keypoint extraction, bundle adjustment utilizing multiple cameras and transorm integration. Firstly, the data association module assign sparse depth observations from LiDAR to image pixels. Subsequently, the direct pose estimation module utilizes the information from all surround-view cameras to estimate the pose of current body frame relative to the latest keyframe. If the current frame is selected as a keyframe, the new keypoints are extracted from it, and the bundle adjuestment for mutiple cameras is performed to further ensure the accuracy of pose estimation. If the current frame is not selected as a keyframe, the relative pose of the current frame and the pose of the latest keyframe are integrated through the trasform integration module, to obtain the final output pose. The whole system is expanded from the monocular DSO \cite{engel2017direct}, while the new keypoint extraction module and the transform integration module are the same as \cite{engel2017direct}. Furthermore, the implementation of data association module is exactly the same as our previous work SDV-LOAM \cite{yuan2023sdv}. Therefore, we only detail the modules highlighted in yellow in Sec. \ref{System Details}.

\section{System Details}
\label{System Details}

\begin{figure}
	\begin{center}
		\includegraphics[width=1\linewidth]{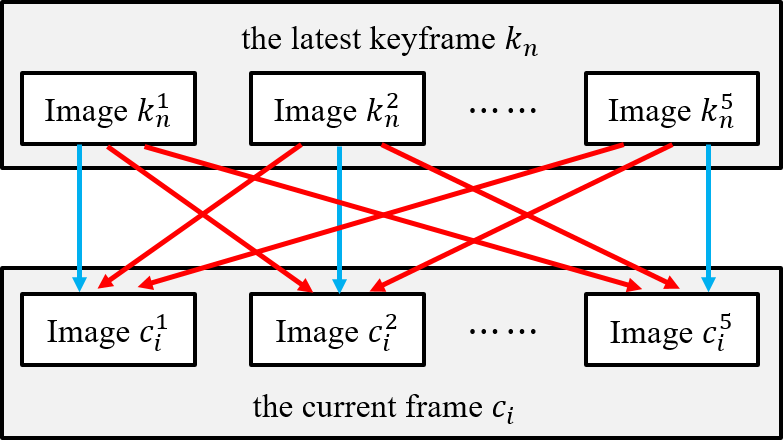}
		\caption{Illustration of ditrect pose estimation utilizing multiple cameras. Each image in the latest keyframe $k_n$, in addition to establishing constraints with the image from the same view of the current frame $c_i$ (as indicated by the blue arrows), can also construct constraints with images from different views within $c_i$ (as indicated by the red arrows).	
		}
		\label{fig3}
	\end{center}
\end{figure}

\subsection{Direct Pose Estimation utilizing Multiple Cameras}
\label{Direct Pose estimation utilizing Multiple cameras}

When the current frame $c_i$ arrives, we track the current frame $c_{i}$ with respect to the newest keyframe $k_{n}$ by projecting extracted keypoints of $k_{n}$ (i.e., $P_{k_{n}^{1}} \sim P_{k_{n}^{5}}$) to $c_{i}$, and then calculate the photometric error between $k_{n}$ and $c_{i}$. Unlike monocular camera frames, each panoramic frame consists of five surround-view images with limited horizontal resolutions. To prevent the issue of insufficient available information due to minimal overlap between the current frame and the latest keyframe from the same view, we allow images from different views to construct photometric errors (as illustrated in Fig \ref{fig3}):
\begin{equation}
	\label{equation3}
	E=\sum_{j=1}^{5} \sum_{l=1}^{5} \sum_{\mathbf{u} \in P_{k_{n}^{j}} }\left\|I_{k_{n}^{j}}(\mathbf{u})-I_{c_{i}^{l}}(\mathbf{u}^{\prime})\right\|_\gamma
\end{equation}
where $\|\cdot\|_{\gamma}$ is the Huber norm and $I(\cdot)$ represents pixel intensity. $\mathbf{u}^{\prime}$ is the projection of $\mathbf{u}$ in $c_{i}^{l}$ calculated as:
\begin{equation}
	\label{equation4}
	\mathbf{u}^{\prime}=\pi\left(\mathbf{T}_{k_{n}^{j}}^{c_{i}^{l}} \pi^{-1}\left(\mathbf{u}, d_{\mathbf{u}}\right)\right)
\end{equation}
Eq. \ref{equation3} involves a total of 25 pose variables (i.e., $\mathbf{T}_{k_n^j}^{c_i^l}$, $j, l=\{1,2,3,4,5\}$) to be optimized. In actual odometry, we aim to estimate the pose of the current body frame relative to the latest body keyframe (i.e., $\mathbf{T}_{k_n^1}^{c_i^1}$), while other pose variables can be expressed as the functional of $\mathbf{T}_{k_n^1}^{c_i^1}$ and camera extrinsic parameters:
\begin{equation}
	\label{equation5}
	\mathbf{T}_{k_n^j}^{c_i^l} = \mathbf{T}_{c_1}^{c_l} \mathbf{T}_{k_n^1}^{c_i^1} \mathbf{T}_{k_j}^{k_1}
\end{equation}

\subsection{Bundle Adjustment utilizing Multiple Cameras}
\label{Bundle Adjustment utilizing Multiple Cameras}

\begin{figure}
	\begin{center}
		\includegraphics[width=1\linewidth]{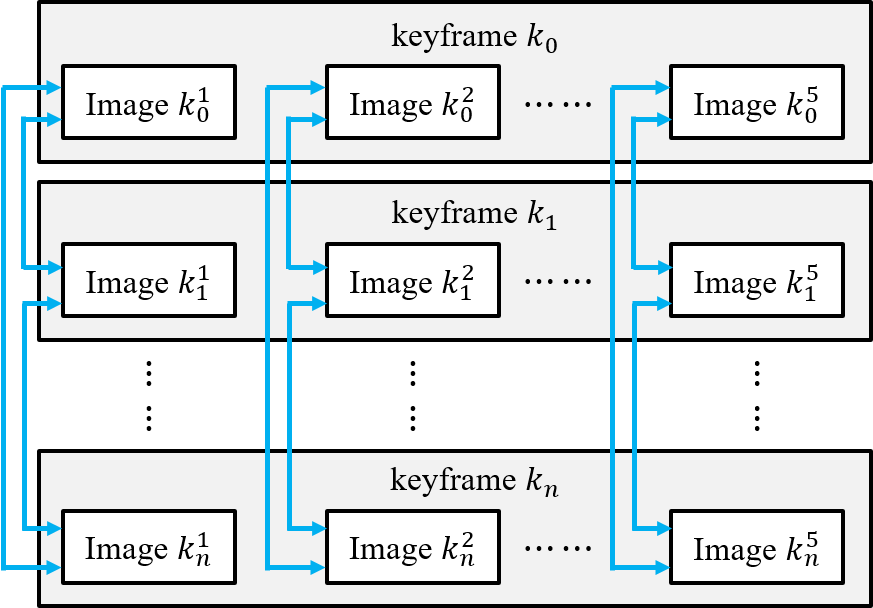}
		\caption{Illustration of bundle adjustment utilizing multiple cameras. Each image in a specifica keyframe constructs constraints with the images from the same view of other keyframes (as indicated by the blue bidirectional arrows).	
		}
		\label{fig4}
	\end{center}
\end{figure}

Sliding window-based bundle adjustment is used to refine the poses of keyframes to reduce accumulative errors in pose estimation. Given that bundle adjustment involves more frames (5$\sim$7 keyframes) than direct pose estimation, we only consider constraints between the images of keyframes from the same view to avoid high computation overhead:
\begin{equation}
	\label{equation6}
	E=\sum_{\substack{s \in \mathcal{F} \\ s \neq l}} \sum_{l \in \mathcal{F}} \sum_{j=1}^{5} \sum_{\mathbf{u} \in P_{k_{s}^{j}}} \left\|I_{k_{s}^{j}}(\mathbf{u})-I_{k_{l}^{j}}(\mathbf{u}^{\prime})\right\|_\gamma
\end{equation}
where $s$, $l$ run over the set of keyframes $\mathcal{F}$ in sliding window. $\mathbf{u}^{\prime}$ is the projection of $\mathbf{u}$ in $k_{l}^{j}$ calculated as:
\begin{equation}
	\label{equation7}
	\mathbf{u}^{\prime}=\pi\left(\mathbf{T}_{k_{s}^{j}}^{k_{l}^{j}} \pi^{-1}\left(\mathbf{u}, d_{\mathbf{u}}\right)\right)
\end{equation}
Eq. \ref{equation6} involves a total of 5$\times n$ pose variables (i.e., $n$ indicates the number of keyframes in sliding window) to be optimized. During the bundle adjustment process, we estimate the body poses of all keyfranes, while other pose variables can be expressed as the function of $\mathbf{T}_{k_s^1}^{w}$ and camera extrinsic parameters:
\begin{equation}
	\label{equation8}
	\mathbf{T}_{k_s^l}^{w} = \mathbf{T}_{k_s^1}^{w} \mathbf{T}_{k_s^l}^{k_s^1}
\end{equation}
Similar as \cite{engel2017direct}, we utilize marginalization to alleviate the computational burden of windowed optimization while retaining previous information.

\begin{table}[]
\begin{center}
	\caption{Details of All Sequences for Evaluation}
	\label{table1}
	\begin{threeparttable}
	\begin{tabular}{p{1.5cm}<{\centering}|p{1.5cm}<{\centering}|p{1.5cm}<{\centering}p{1.5cm}<{\centering}}
		\hline
		Dataset               & Sequence   & Time      & Weather \\ \hline
		\multirow{6}{*}{$nclt$} & 2012-01-08 & Midday    & Cloudy  \\
		& 2012-09-28 & Evening   & Sunny   \\
		& 2012-11-04 & Morning   & Cloudy  \\
		& 2012-12-01 & Evening   & Sunny   \\
		& 2013-02-23 & Afternoon & Cloudy  \\
		& 2013-04-05 & Afternoon & Sunny   \\ \hline
		\multirow{2}{*}{$ijrr$} & ford\_1    & -         & -       \\
		& ford\_2    & -         & -       \\ \hline
	\end{tabular}
	\end{threeparttable}
	\begin{tablenotes}
	\footnotesize
	\item[] \textbf{Denotations}: "-" means the corresponding content is not available.
	\end{tablenotes}
\end{center}
\end{table}

\section{Experiments}
\label{Experiments}

In this section, we evaluate our panoramic direct LiDAR-assisted VO on two public datasets: $nclt$ \cite{carlevaris2016university} and $ijrr$ \cite{pandey2011ford}. $nclt$ dataset consists of 27 sequences which are logged with sensors mounted on top of a Segway in driving scenarios. The Segway is equipped with a Ladybug3 panoramic camera, a Velodyne HDL-32E LiDAR, a real-time kinematic (RTK) global positioning system (GPS), etc. Our system takes data from the Velodyne HDL-32E LiDAR and the Ladybug3 panoramic camera as input. For every panoramic image from the camera, we use the distortion parameters provided by \cite{carlevaris2016university} to perform distortion correction. The Segway repeatedly explore the campus, both indoors and outdoors, on varying trajectories, at different time and weather. $ijrr$ dataset consists of 2 sequences which are logged with sensors mounted on top of a vehicle in driving scenarios. The vehicle is equipped with a Ladybug3 panoramic camera, a Velodyne HDL-64E LiDAR, a Inertial Navigation System(INS) with GPS, etc. Our system takes data from the Velodyne HDL-64E LiDAR and the Ladybug3 panoramic camera as input. For every panoramic image from the camera, we use the distortion parameters provided by \cite{pandey2011ford} to perform distortion correction.

We select 6 representative sequences including different time and weather from $nclt$, and all sequences from $ijrr$ for evaluation. The details of all the 8 sequences, indluding time and weather, are listed in Table \ref{table1}. Since the $ijrr$ dataset does not explicitly state the time and weather involved in sequences, we do not record them in Table \ref{table1}.

In sequences of $nclt$, the segment of traversing the doorway, is usually accompanied by drastic changes in illumination, which poses a significant challenge for LiDAR-assisted VO. Especially for the panoramic images with low frame rates (i.e., 5 fps), the rapid transition between indoor and outdoor environments results in abrupt changes in illumination, and cause the VO to lose tracking. Consequently, we manually segment the selected sequences as delineated in \cite{jiang2021panoramic}. The outdoor segments are utilized to demonstrate the outstanding performance of our system in expansive environments, while the indoor segments are employed to demonstrate the great performance of our system performance within small-scale scenes.

\begin{figure}
	\begin{center}
		\includegraphics[width=1\linewidth]{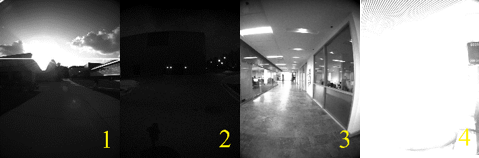}
		\caption{Challenging outdoor and indoor environments of $nclt$ dataset, including overexposed outdoor scenarios (denoted as label 1), underexposed outdoor scenarios (denoted as label 2), reflective indoor scenarios (denoted as label 3) and overexposed indoor scenarios (denoted as label 4).	
		}
		\label{fig5}
	\end{center}
\end{figure}

The selected outdoor segments are representative as they encompass different environmental lighting conditions: the soft sunlight of morning, the intense sunlight of midday, and the dim light of the evening. The indoor segments encompass extreme situations such as reflection and overexposure. Fig \ref{fig5} illustrates a variety of challenging environments in selected segments including overexposed outdoor scenarios, underexposed outdoor scenarios, reflective indoor scenarios and overexposed indoor scenarios.

We use the evaluation tool $evo$ \cite{grupp2017evo} to evaluate the root mean square error (RMSE) of absolute translational error (ATE) between the estimated pose and the ground-truth pose in all experiments. A consumer-level computer equipped with an Intel Core i7-11700 and 32 GB RAM is used for all experiments.

\begin{table*}
	\caption{RMSE of ATE Comparison with State-of-the-Art Methods}
	\label{table2}
	\begin{center}
		\begin{threeparttable}
			\begin{tabular}{p{1.5cm}<{\centering}|p{1.5cm}<{\centering}|p{1.5cm}<{\centering}p{1.5cm}<{\centering}|p{1.5cm}<{\centering}p{1.5cm}<{\centering}p{1.5cm}<{\centering}|p{1.5cm}<{\centering}|p{1.5cm}<{\centering}}
				\hline
				Sequences                                                                                       & Segment & \begin{tabular}[c]{@{}c@{}}Length of\\ Segment (m)\end{tabular} & \begin{tabular}[c]{@{}c@{}}Number \\ of Frames\end{tabular} & LIMO  & PAN-SLAM & SDV-LOAM & \begin{tabular}[c]{@{}c@{}}Ours\\ (Mono)\end{tabular} & Ours  \\ \hline
				\multirow{6}{*}{\begin{tabular}[c]{@{}c@{}}2012-01-08\\ Midday\\ Cloudy\end{tabular}}    & 1-1     & 3535                                                            & 14036                                                       & 35.48 & 13.62    & 68.61        & 21.65                                                 & \textbf{4.77}  \\
				& 1-2     & 270                                                             & 1046                                                        & 1.64  & 38.18    & 8.50        & 1.62                                                  & \textbf{0.39}  \\
				& 1-3     & 221                                                             & 805                                                         & 5.97  & 0.22     & 4.18        & 10.90                                                 & \textbf{0.13}  \\
				& 1-4     & 41                                                              & 421                                                         & 2.16  & \textbf{0.10}     & 6.48        & 0.26                                                  & 0.11  \\
				& 1-5     & 1010                                                            & 3833                                                        & 9.01  & 3.20     & 50.46        & 49.43                                                 & \textbf{1.32}  \\
				& 1-6     & 566                                                             & 2272                                                        & 1.95  & 1.90     & 21.97        & 2.50                                                  & \textbf{1.00}  \\ \hline
				\multirow{5}{*}{\begin{tabular}[c]{@{}c@{}}2012-09-28\\ Evening\\ Sunny\end{tabular}}    & 2-1     & 2726                                                            & 8902                                                        & 21.55 & 43.03    & $\times$        & 24.00                                                 & \textbf{17.20} \\
				& 2-2     & 1034                                                            & 4076                                                        & 9.60  & 6.01     & 46.32        & 11.61                                                 & \textbf{0.91}  \\
				& 2-3     & 366                                                             & 1432                                                        & 17.35 & $\times$        & 14.30        & 0.43                                                  & \textbf{0.34}  \\
				& 2-4     & 998                                                             & 4311                                                        & $\times$     & 18.10    & $\times$        & 51.88                                                 & \textbf{4.32}  \\
				& 2-5     & 376                                                             & 1512                                                        & 42.69 & 1.01     & 26.52        & 51.92                                                 & \textbf{0.50}  \\ \hline
				\multirow{5}{*}{\begin{tabular}[c]{@{}c@{}}2012-11-04\\ Morning\\ Cloudy\end{tabular}}   & 3-1     & 525                                                             & 3097                                                        & 32.14 & 14.42    & 33.72        & 5.19                                                  & \textbf{0.73}  \\
				& 3-2     & 1009                                                            & 4475                                                        & 34.40 & 2.88     & $\times$        & 9.06                                                  & \textbf{1.70}  \\
				& 3-3     & 518                                                             & 3043                                                        & 2.90  & \textbf{0.46}     & 24.78        & 8.91                                                  & 0.74  \\
				& 3-4     & 952                                                             & 3905                                                        & 24.77 & 54.68    & 83.76        & 14.97                                                 & \textbf{2.26}  \\
				& 3-5     & 1122                                                            & 5279                                                        & \textbf{1.61}  & 16.21    & 50.03        & 15.06                                                 & 2.95  \\ \hline
				\multirow{5}{*}{\begin{tabular}[c]{@{}c@{}}2012-12-01\\ Evening\\ Sunny\end{tabular}}    & 4-1     & 408                                                             & 2597                                                        & 16.97 & 2.97     & 14.69        & 1.99                                                  & \textbf{0.67}  \\
				& 4-2     & 963                                                             & 4112                                                        & 6.55  & $\times$        & 54.28        & 27.78                                                 & \textbf{1.96}  \\
				& 4-3     & 1648                                                            & 6993                                                        & 38.08 & 45.41    & 48.42        & 10.12                                                 & \textbf{5.88}  \\
				& 4-4     & 229                                                             & 942                                                         & 2.72  & 0.34     & 23.91        & 1.10                                                  & \textbf{0.33}  \\
				& 4-5     & 1114                                                            & 4896                                                        & $\times$     & $\times$        & 35.43        & 29.31                                                 & \textbf{22.92} \\ \hline
				\multirow{5}{*}{\begin{tabular}[c]{@{}c@{}}2013-02-23\\ Afternoon\\ Cloudy\end{tabular}} & 5-1     & 582                                                             & 2820                                                        & 1.22  & 1.78     & 42.34        & 1.25                                                  & \textbf{0.74}  \\
				& 5-2     & 645                                                             & 2573                                                        & 7.12  & 3.89     & 28.92        & \textbf{1.08}                                                  & 1.41  \\
				& 5-3     & 621                                                             & 2616                                                        & 4.51  & $\times$        & 24.12        & 0.70                                                  & \textbf{0.67}  \\
				& 5-4     & 2687                                                            & 11735                                                       & \textbf{5.93}  & 17.96    & $\times$        & $\times$                                                 & 9.27  \\
				& 5-5     & 241                                                             & 1000                                                        & 23.72 & 0.71     & 23.46        & 0.69                                                  & \textbf{0.37}  \\ \hline
				\multirow{4}{*}{\begin{tabular}[c]{@{}c@{}}2013-04-05\\ Afternoon\\ Sunny\end{tabular}}  & 6-1     & 430                                                             & 2728                                                        & 15.11 & 63.45    & $\times$        & 1.40                                                  & \textbf{0.93}  \\
				& 6-2     & 1649                                                            & 7221                                                        & 3.06  & \textbf{2.36}     & $\times$        & 19.58                                                 & 3.09  \\
				& 6-3     & 2153                                                            & 5641                                                        & 21.42 & 7.94     & $\times$        & $\times$                                                 & \textbf{3.40}  \\
				& 6-4     & 391                                                             & 1617                                                        & 15.39 & $\times$        & 18.35        & 1.01                                                  & \textbf{0.68}  \\ \hline
				\multirow{5}{*}{\begin{tabular}[c]{@{}c@{}}2012-01-08\\ Midday\\ Indoor\end{tabular}}    & 7-1     & 237                                                             & 1459                                                        & $\times$     & 0.57     & 4.53        & 0.57                                                  & \textbf{0.53}  \\
				& 7-2     & 85                                                              & 614                                                         & $\times$     & 0.40     & 4.39        & 0.51                                                  & \textbf{0.31}  \\
				& 7-3     & 160                                                             & 1069                                                        & $\times$     & 0.32     & 6.13        & 1.46                                                  & \textbf{0.24}  \\
				& 7-4     & 124                                                             & 717                                                         & $\times$     & \textbf{0.18}     & 3.68        & 1.26                                                  & 0.28  \\
				& 7-5     & 50                                                              & 414                                                         & $\times$     & 0.13     & 2.52        & \textbf{0.11}                                                  & 0.24  \\ \hline
				\multirow{4}{*}{\begin{tabular}[c]{@{}c@{}}2012-09-28\\ Evening\\ Indoor\end{tabular}}   & 8-1     & 108                                                             & 498                                                         & $\times$     & 0.55     & 5.32        & 0.25                                                  & \textbf{0.19}  \\
				& 8-2     & 182                                                             & 910                                                         & $\times$     & 0.46     & 3.76        & 0.45                                                  & \textbf{0.44}  \\
				& 8-3     & 75                                                              & 333                                                         & $\times$     & 3.28     & 1.70        & 0.12                                                  & \textbf{0.11}  \\
				& 8-4     & 38                                                              & 192                                                         & $\times$     & 2.24     & 0.41        & \textbf{0.08}                                                  & 0.10  \\ \hline
				\multirow{3}{*}{\begin{tabular}[c]{@{}c@{}}2012-11-04\\ Evening\\ Indoor\end{tabular}}   & 9-1     & 40                                                              & 287                                                         & $\times$     & \textbf{0.09}     & 1.06        & 0.14                                                  & 0.10  \\
				& 9-2     & 145                                                             & 677                                                         & $\times$     & \textbf{0.38}     & 11.45        & 1.47                                                  & 0.41  \\
				& 9-3     & 1287                                                            & 6020                                                        & $\times$     & 59.30    & $\times$        & 36.21                                                 & \textbf{7.53}  \\ \hline
				\begin{tabular}[c]{@{}c@{}} ford\_1\end{tabular}                                          & 10-1    & 1500                                                            & 2762                                                        & 9.73  & 13.20    & 30.36        & 40.38                                                 & \textbf{2.80}  \\ \hline
				\multirow{2}{*}{\begin{tabular}[c]{@{}c@{}} ford\_2\end{tabular}}                         & 11-1    & 1319                                                            & 1460                                                        & $\times$     & $\times$        & $\times$        & $\times$                                                 & \textbf{2.26}  \\
				& 11-2    & 1219                                                            & 1519                                                        & $\times$     & 81.07    & $\times$        & 49.74                                                 & \textbf{2.41}  \\ \hline
			\end{tabular}
		\end{threeparttable}
		\begin{tablenotes}
			\footnotesize
			\item[] \textbf{Denotations}: As all approaches fail to run entirely on selected sequences, we disassemble each sequence into segments and evaluate on them. "$\times$" means the approach fail to run entirely on the corresponding segment.
		\end{tablenotes}
	\end{center}
\end{table*}

\subsection{Comparison with State-of-the-Arts}
\label{Comparison with State-of-the-Arts}

We compare our panoramic direct LiDAR-assisted VO with three state-of-the-art approaches, i.e., LIMO \cite{graeter2018limo}, PAN-SLAM \cite{ji2020panoramic} and SDV-LOAM \cite{yuan2023sdv}, on all testing sequences. Among them, LIMO is a LiDAR-assisted monocular VO which involves semantic information as input. Since both $nclt$ and $ijrr$ do not provide semantic labels, we deactivate the relevant parts of handling semantic information in LIMO. Meanwhile, LIMO requires the height value of the LiDAR above ground as the input parameter, which is inferred from the extrinsic parameters provided by the two datasets. PAN-SLAM is a panoramic VO without the assistance of LiDAR, and we disable loop-closing of PAN-SLAM for eliminating the interference of the loop-closure module. SDV-LOAM is formed by cascading a monocular semi-direct LiDAR-assisted VO and a LiDAR odometry, and we disable the LiDAR odometry module. Both LIMO and SDV-LOAM are designed for monocular camera and 3D LiDAR, so we utilize the images from the forward-facing camera of panoramic camera and the 3D points from LiDAR as input for them.

There are few methods that we can compare to on these challenging sequences. Although we attempted to run DSO \cite{engel2017direct} and DEMO \cite{jin2013demo} on these sequences, both two methods failed on most of selected segments. The major reason is that both two methods are monocular VO or LiDAR-assisted VO, whose robustness is severely compromised when poor image quality. Another category of vision-LiDAR odometry, such as V-LOAM \cite{zhang2015visual} and SR-LIVO \cite{yuan2022sr, yuan2024sr}, primarily utilize 3D LiDAR points for pose estimation. However, we focus on exploring the bottlenecks of existing LiDAR-assisted VO in this work, and thus do not compare with them.

Table \ref{table2} records the pose estimation results of all testing approaches and our panoramic direct LiDAR-assisted VO. Due to the challenging environment for both VO and LiDAR-assisted VO, no approach is able to run entirely on any selected sequences. Therefore, we disassemble each sequence into sevral segments and evaluate the testing approaches on these segments. Results in Table \ref{table2} demonstrate that our panoramic direct LiDAR-assisted VO outperforms visual module outperforms all testing approaches on most segments in terms of smaller RMSE of ATE. PAN-SLAM fails to run entirely on 6 segments and achieves large ATE results on 7 sequences. This indicates that even with the full utilization of a 360-degree panoramic image, the performance of VO remains fragile in challenging scenarios if lacking depth sensors to provide distance information. In contrast, our panoramic LiDAR-assisted VO achieves smaller ATE across almost all testing segments, demonstrating the advantage of integrating panoramic camera with 3D LiDAR. On the other hand, LIMO fails to run entirely on 15 sequences and achieves large ATE results on 5 sequeces, while SDV-LOAM fails to run entirely on 10 sequences and achieves large ATE results on 10 sequences. This result indicates that even with the distance information perceived by 3D LiDAR, VO still cannot run stably and robustly if the available information from images are insufficient.

\subsection{Impact of Panoramic FOV for Our System}
\label{Impact of Panoramic FOV for Our System}

To demonstrate the significant improvement of utilizing panoramic FOV for LiDAR-assisted VO, we design this ablation study by using measurements from only the forward-facing camera and 3D LiDAR for pose estimation while remaining all other system parameters unchanged. Results in Table \ref{table2} demonstrate that the accuracy and robustness of utilizing panoramic images signicantly outperforms utilizing only forward-facing images. The major reason is that the available information from one camera is limited and unstable. In comparison, the panoramic image can provide exploitable information from other views when a single-view image is subject to environmental interference such as overexposure or motion blur, thereby significantly enhancing the accuracy and robustness of LiDAR-assisted VO.

\subsection{Impact of Constraints from Different Views}
\label{Impact of Constraints from Different Views}

\begin{table}[]
\begin{center}
	\caption{Ablation Study of Constraints from Different Views}
	\label{table3}
	\begin{threeparttable}
	\begin{tabular}{p{1.5cm}<{\centering}|p{1.5cm}<{\centering}|p{1.5cm}<{\centering}|p{1.5cm}<{\centering}}
		\hline
		Sequences                                                                              & Segment & \begin{tabular}[c]{@{}c@{}}Ours\\ w/o CFDV\end{tabular} & Ours          \\ \hline
		\multirow{5}{*}{\begin{tabular}[c]{@{}c@{}}2012-11-04\\ Morning\\ Cloudy\end{tabular}} & 3-1     & 1.05                                                    & \textbf{0.73} \\
		& 3-2     & 1.83                                                    & \textbf{1.71} \\
		& 3-3     & 1.92                                                    & \textbf{0.74} \\
		& 3-4     & \textbf{1.71}                                           & 2.26          \\
		& 3-5     & 3.12                                                    & \textbf{2.95} \\ \hline
		\multirow{3}{*}{\begin{tabular}[c]{@{}c@{}}2012-11-04\\ Evening\\ Indoor\end{tabular}} & 9-1     & \textbf{0.08}                                           & 0.10          \\
		& 9-2     & 0.68                                                    & \textbf{0.41} \\
		& 9-3     & 10.50                                                   & \textbf{7.53} \\ \hline
	\end{tabular}
	\end{threeparttable}
	\begin{tablenotes}
	\footnotesize
	\item[] \textbf{Denotations}: "CFDV" is the abbreviation of "Constraints from Different Views".
	\end{tablenotes}
\end{center}
\end{table}

As mentioned in Sec. \ref{Direct Pose estimation utilizing Multiple cameras}, constructing phomometric constraints from different views can take full advantage of more available information from panoramic images, and in turn improve the performance of pose estimation. In this section, we design the ablation study of estimating poses with vs. without constraints from different views. Results in Table \ref{table3} demonstrate that utilizing constraints from different views can improve the accuracy of pose estimation on most segments of sequence 2012-11-04. Due to the limitation of page number, we only present the results on one sequence, and the other sequences yielded similar results.

\subsection{Visualization}
\label{Visualization}

\begin{figure}
	\begin{center}
		\includegraphics[scale=0.4]{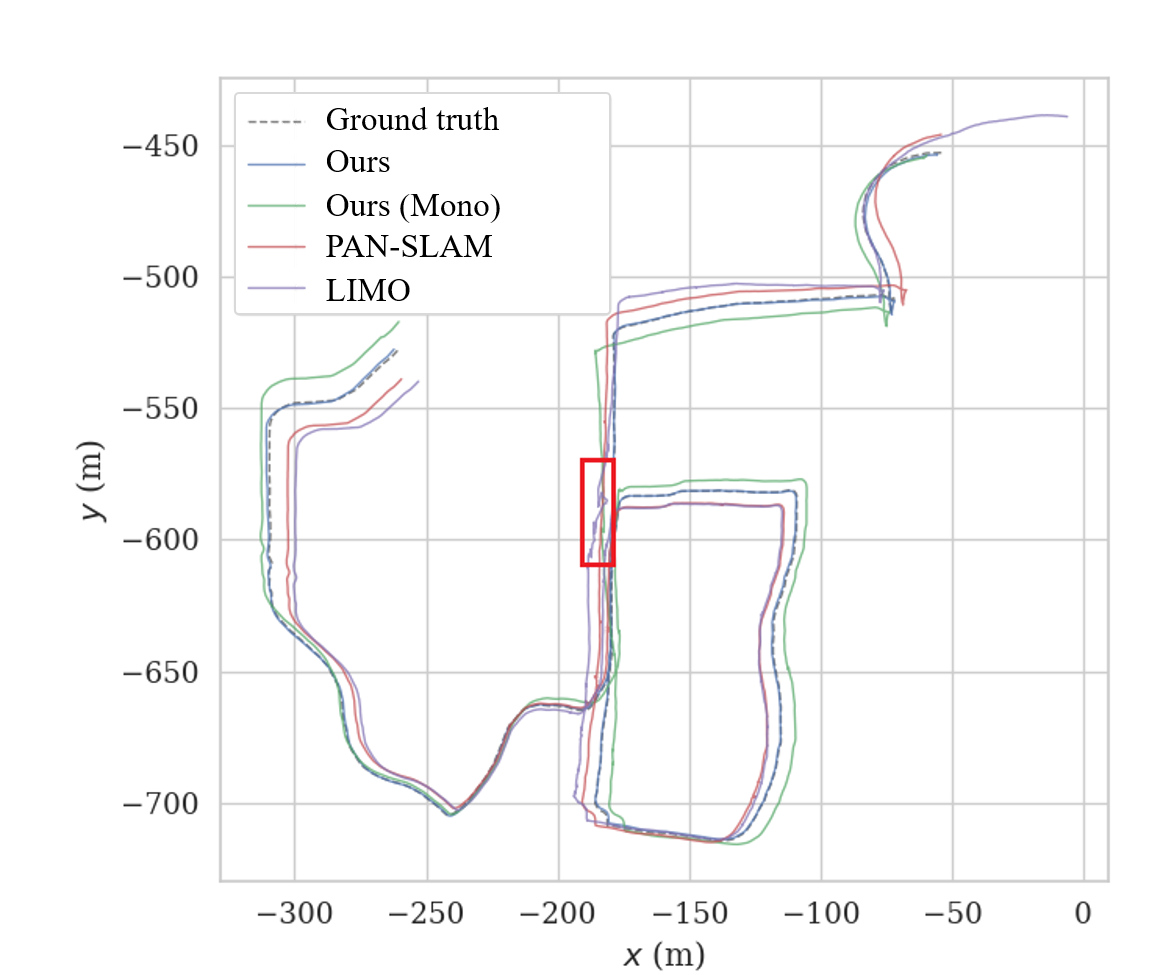}
		\caption{Trajectories of segment 2-2. The dotted line, blue line, green line, red line and purple line represents ground truth, our system, our system utilizing only single-view images, PAN-SLAM and LIMO respectively. The trajectory estimated by our system is almost perfectly overlaid with the ground truth. The location experiencing excessive exposure in the forward-facing view (denoted as lable 1 in Fig. \ref{fig5}) corresponds to the area delineated by the red rectangle. LIMO and our system utilizing only single-view images, which solely rely on the foreground images for pose estimation, encounters pronounced trajectory disorientation at this situation.
		}
		\label{fig6}
	\end{center}
\end{figure}

Fig.\ref{fig6} compares the estimated trajectories of almost all testing approaches with ground truth on segment 2-2. The reason for not plotting the trajectory of SDV-LOAM is that its estimated trajectory on this segment is too poor to provide any references. In the remaining visualization comparison results, the trajectory estimated by our system is almost entirely overlaid with the ground truth, which further demonstrates the superiority of our approach. When the utilized image is reduced from surround panoramic view to solely forward-facing view, the accuracy of estimated trajectory significantly deteriorates. The location experiencing excessive exposure in the forward-facing view (denoted as lable 1 in Fig. \ref{fig5}) corresponds to the area delineated by the red rectangle in Fig. \ref{fig6}. LIMO and our system utilizing only single-view images, which solely rely on the foreground images for pose estimation, encounters pronounced trajectory disorientation at this situation. In comparison, our system utilizing panoramic-view images and PAN-SLAM maintain a smooth trajectory even in such challenging situations, further demonstrating the significant role of panoramic-view in enhancing the robustness of VO and LiDAR-assisted VO.

\section{Conclusion}
\label{Conclusion}

In this paper, we propose a panoramic direct LiDAR-assisted visual odometry system for accurate and robust pose estimation in outdoor environments. The system is extended from monocular DSO \cite{engel2017direct}, while the LiDAR observation and the panoramic image processing module are integrated into \cite{engel2017direct}. To address the issue of limited overlap between consecutive frames form a single-view, our system utilize the images from different views to construct constraints in pose estimation. Experimental results demonstrate the benefit of panoramic images for LiDAR assisted VO and the effectiveness of constructing constraints between different views. Future work includes integrating IMU into our system.

\bibliographystyle{IEEEtran}
\bibliography{IEEEabrv,IEEEExample}

\end{document}